\documentclass[conference,a4paper]{IEEEtran}

\usepackage{times}
\usepackage{epsfig}
\usepackage{graphicx}
\usepackage{amsmath}
\usepackage{amssymb}

\usepackage{amsfonts}
\usepackage{color}
\usepackage{url}

\newcommand{\ie}{{\it i.e.}}
\newcommand{\eg}{{\it e.g.}}
\newcommand{\etal}{{\it et al.}}

\DeclareMathOperator{\sgn}{sgn}

\begin{document}
\title{One-Class Slab Support Vector Machine}

\author{Victor Fragoso$^{\dagger}$ ~~~~ Walter Scheirer$^{\ddagger}$ ~~~~ Joao Hespanha$^{\mp}$ ~~~~ Matthew Turk$^{\mp}$ \\
$^\dagger$West Virginia University  ~~ $^\ddagger$University of Notre Dame ~~ $^\mp$University of California, Santa Barbara \\
\small{victor.fragoso@mail.wvu.edu ~~ wscheire@nd.edu ~~ \{hespanha@ece, mturk@cs\}.ucsb.edu}
}

\maketitle

\begin{abstract}
This work introduces the one-class slab SVM (OCSSVM), a one-class classifier that aims at improving the performance of the one-class SVM. The proposed strategy reduces the false positive rate and increases the accuracy of detecting instances from novel classes. To this end, it uses two parallel hyperplanes to learn the normal region of the decision scores of the target class. OCSSVM extends one-class SVM since it can scale and learn non-linear decision functions via kernel methods. The experiments on two publicly available datasets show that OCSSVM can consistently outperform the one-class SVM and perform comparable to or better than other state-of-the-art one-class classifiers. 
\end{abstract}

\IEEEpeerreviewmaketitle

\section{Introduction}
\vspace{-2mm}

\begin{figure}[t]
\centering
\includegraphics[width=0.5\textwidth]{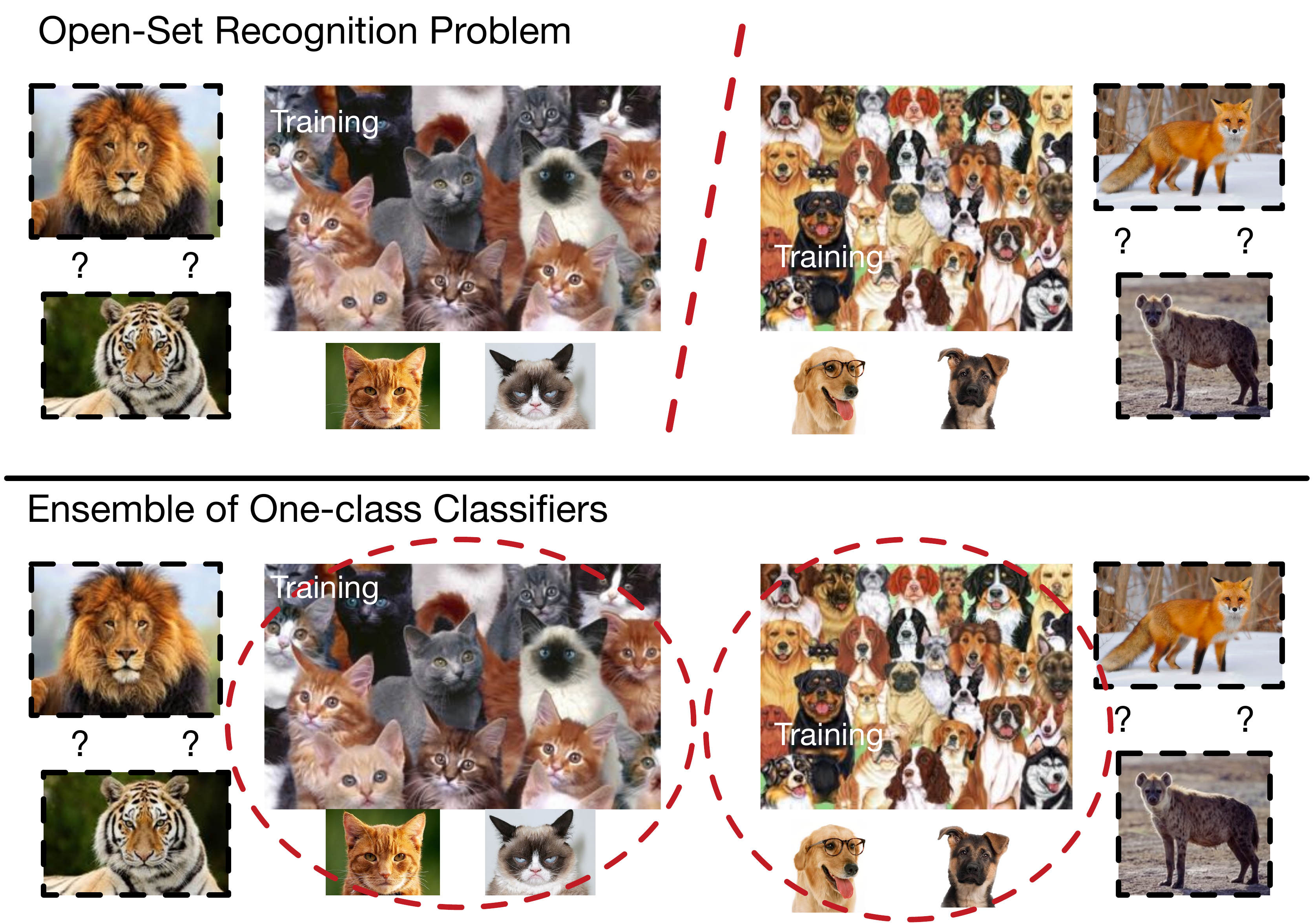}
\vspace{-6mm}
\caption{The open-set recognition problem (top) challenges existing recognition systems. This is because classifiers can face instances from novel or unknown classes (images with dashed-frames). These novel classes cause failures during prediction time. Collecting instances from all the possible classes is a challenging task in many applications. For instance, collecting and labeling instances of all existing animals to avoid this problem is impractical. An ideal solution to this open-set recognition problem is an ensemble of one-class classifiers (bottom). A single one-class classifier only requires instances of a target positive class to train (illustrated as circles). Such classifiers detect samples from the target classes and identify unknown instances. However, their performance needs improvement in order to solve the open-set recognition problem. The proposed approach improves the performance of the one-class SVM. It is a step towards the solution of the open-set recognition problem with an ensemble of one-class classifiers.}
\label{fig:open_set}
\vspace{-4mm}
\end{figure}

Current recognition systems perform well when their training phase uses a vast amount of samples from all classes encountered at test time. However, these systems significantly decrease in performance when they face the open-set recognition problem~\cite{Scheirer_2013_TPAMI}: recognition in the presence of samples from unknown or novel classes. This occurs even for already solved datasets (\eg, the Letter dataset~\cite{frey91}) that are recontextualized as open-set recognition problems. The top of the Figure~\ref{fig:open_set} illustrates the general open-set recognition problem.

Recent work has aimed at increasing the robustness of classifiers in this context~\cite{Bendale_2015_CVPR, Scheirer_2014_TPAMIb, Scheirer_2013_TPAMI}. However, these approaches assume knowledge of at least a few classes during the training phase. Unfortunately, many recognition systems only have a few samples from just the target class. For example, collecting images from the normal state of a retina is easier than collecting those from abnormal retinas~\cite{yamuna2013detection}. 

One-class classifiers are useful in applications where collecting samples from negative classes is challenging, but gathering instances from a target class is easy. An ensemble of one-class classifiers can solve the open-set recognition problem. This is because each one-class classifier can recognize samples of the class it was trained for and detect novel samples; see Figure~\ref{fig:open_set} for an illustration of the ensemble of one-class classifiers. Unlike other solutions to the open-set recognition problem (\eg, the 1-vs-Set SVM~\cite{Scheirer_2013_TPAMI}), the ensemble offers  parallelizable training and easy integration of new categories.
These computational advantages follow from the independence of each classifier and allow the ensemble to scale well with the number of target classes.

However, the one-class classification problem is a challenging binary categorization task. This is because the classifier is trained with only positive examples from the target class, yet, it must be able to detect novel samples (negative class data). For instance, a one-class classifier trained to detect normal retinas must learn properties from them to recognize other images of normal and abnormal retinas. A vast amount of research has focused on tackling the challenges faced in the one-class classification problem. These strategies include statistical methods~\cite{clifton2011novelty, ritter1997outliers}, neural networks~\cite{bishop1994novelty, Moya1996463}, and kernel methods~\cite{hoffmann2007kernel, scholkopf_ocsvm, tax2004support}.

Despite the advancements, the performance of one-class classifiers falls short for open-set recognition problems. To improve the performance of one-class classifiers, we propose a new algorithm called the one-class slab SVM (OCSSVM), which reduces the rate of classifying instances from a novel class as positive (false positive rate) and increases the rate of detecting instances from a novel class (true negative rate). This work focuses on the one-class SVM classifier as a basis because it can scale well and can learn non-linear decision functions via kernel methods. 


\begin{figure*}[t]
\centering
\includegraphics[width=0.9\textwidth]{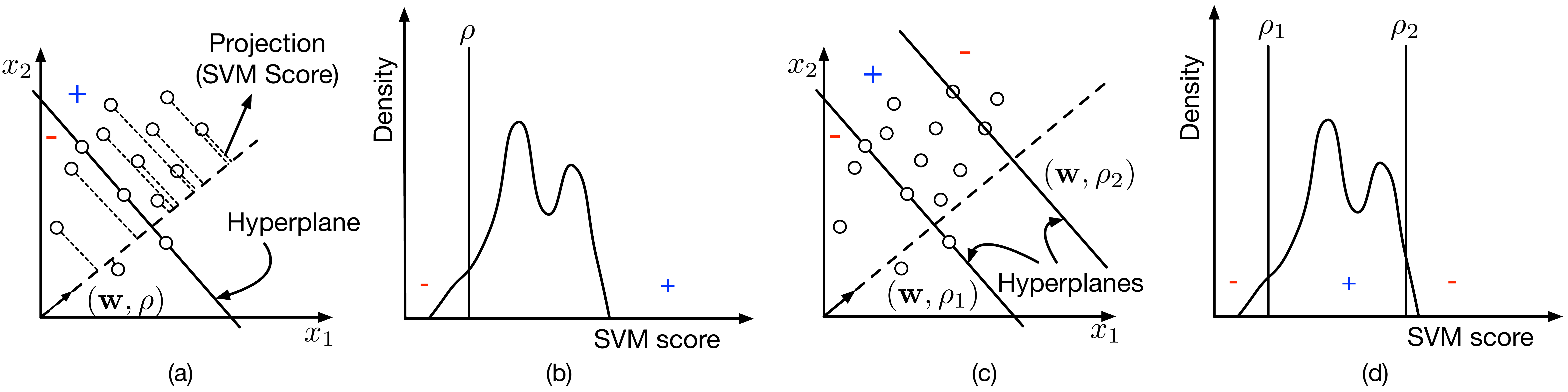}
\vspace{-4mm}
\caption{ {\bf(a)} The one-class SVM (OCSVM) learns a hyperplane that softly maximizes the margin between the origin and the data from the positive class. Its decision function projects the data onto the normal vector $\mathbf{w}$ to produce the SVM scores. {\bf(b)} Subsequently, the decision function labels the samples as negative when the SVM scores fall below a threshold $\rho$, or labels them as positive otherwise. However, the one-class SVM does not account for outliers that can occur on the right tail of the SVM score density. In this case, a high rate of false positives can occur. {\bf(c)} The proposed strategy considers learning two hyperplanes with the same normal vector but with different offsets. {\bf (d)} These hyperplanes learn the ``normal'' region for the SVM scores. This region is called a slab.}
\label{fig:ocsvm_intuition}
\vspace{-4mm}
\end{figure*}

The one-class SVM (OCSVM) learns a hyperplane that keeps most of the instances of the target class on its positive side. However, instances from the negative class can also be on the positive side of this hyperplane. The OCSVM does not account for this case, which makes it prone to a high false positive rate. Unlike the OCSVM, the proposed OCSSVM approach encloses the normal region of the target class in feature space by using two parallel hyperplanes. When an instance falls inside the normal region or the slab created by the hyperplanes, the OCSSVM labels it as a sample from the target class, and negative otherwise. Figure~\ref{fig:ocsvm_intuition} provides an overview of this new algorithm.

Using two parallel hyperplanes has been explored before in visual recognition problems. Cevikalp and Triggs~\cite{cevikalp2012efficient} proposed a cascade of classifiers for object detection. Similarly, Scheirer~\etal~\cite{Scheirer_2013_TPAMI} proposed the 1-vs-Set SVM, where a greedy algorithm calculates the slab parameters after training a regular linear SVM. However, these methods are not strictly one-class classifiers since they use samples from known negative classes. Parallel hyperplanes have also been used by Giesen~\etal~\cite{giesen2004kernel} to compress a set of 3D points and by Glazer~\etal~\cite{glazer2013q} to estimate level sets from a high-dimensional distribution. In contrast to these methods, the OCSSVM targets the open-set recognition problem directly and computes the optimal size of the slab automatically.



This work presents two experiments on two publicly available visual recognition datasets. This is because visual recognition systems encounter novel classes very frequently in natural scenes that contain both target and novel objects. The experiments evaluate the performance of the proposed approach and compare it with other state-of-the-art one-class classifiers. The experiments show that OCSSVM consistently outperforms the one-class SVM and performs comparable to or better than other one-class classifiers. 

The OCSSVM represents a step towards the ideal robust recognition system based on an ensemble of one-class classifiers. The proposed OCSSVM can also improve the performance of other applications such as the identification of abnormal episodes in gas-turbines~\cite{clifton2008bayesian}; the detection of abnormal medical states from vital signs~\cite{Hyoung05}; and the detection of impostor patterns in a biometric system~\cite{Hyoung05}.

\vspace{-1mm}
\subsection{Brief Review of the One-class SVM}
\vspace{-1mm}

Sch\"{o}lkopf \etal~\cite{scholkopf_ocsvm} proposed the one-class support vector machine (OCSVM) to detect novel or outlier samples. Their goal was to find a function that returns +1 in a ``small'' region capturing most of the target data points, and -1 elsewhere. Their strategy consists of mapping the data to a feature space via kernel methods. Subsequently, it finds a hyperplane in this new feature space that maximizes the margin between the origin and the data.

To find this hyperplane, Sch\"{o}lkopf \etal~ proposed the following optimization problem:
\begin{equation}
\vspace{-1.5mm}
\begin{aligned}
& \underset{\mathbf{w}, \rho, \boldsymbol{\xi}}{\text{minimize}}
& & \frac{1}{2} \|\mathbf{w}\|^2 + \frac{1}{\nu m}\sum_{i=1}^m \xi_i - \rho \\
& \text{subject to} & & \left\langle\mathbf{w}, \Phi(\mathbf{x}_i) \right\rangle \geq \rho - \xi_i, \\
& & & \xi_i \geq 0, \; i = 1, \ldots, m,
\end{aligned}
\label{eq:ocsvm}
\end{equation}
\noindent where $m$ is the number of total training samples from the target class; $\nu$ is an upper-bound on the fraction of outliers and a lower bound on the fraction of support vectors (SV); $\mathbf{x}_i$ is the $i$-th training sample feature vector; $\mathbf{w}$ is the hyperplane normal vector; $\Phi()$ is a feature map; $\boldsymbol{\xi}$ are slack variables; and $\rho$ is the offset (or threshold). Solvers for this problem compute the dot product $\left\langle \Phi(\mathbf{x}_i), \Phi(\mathbf{x}_j) \right\rangle$ via a kernel function $k(\mathbf{x}_i, \mathbf{x}_j) = \langle \Phi(\mathbf{x}_i), \Phi(\mathbf{x}_j)\rangle$. 

Sch\"{o}lkopf \etal~\cite{scholkopf_ocsvm} proposed to solve the problem shown in Eq.~\eqref{eq:ocsvm} via its dual problem:
\begin{equation}
\vspace{-1mm}
\begin{aligned}
& \underset{\boldsymbol{\alpha}}{\text{minimize}}
& & \frac{1}{2} \boldsymbol{\alpha}^T K \boldsymbol{\alpha} \\
& \text{subject to} & & \| \boldsymbol{\alpha} \|_1 = 1, \\
& & & 0 \leq \alpha_i \leq \frac{1}{\nu m},\; i = 1, \ldots, m,
\end{aligned}
\label{eq:qp_svm_oneclass}
\end{equation}
\noindent where $K$ is the kernel matrix calculated using a kernel function, \ie, $K_{ij} = k(\mathbf{x}_i, \mathbf{x}_j)$, and $\boldsymbol{\alpha}$ are the dual variables. This optimization problem is a constrained quadratic-program which is convex. Thus, solvers can use Newton-like methods~\cite{boyd2004convex, scholkopf2001learning} or a variant of the sequential-minimal-optimization (SMO) technique~\cite{platt_smo}.

The SVM decision function is calculated as follows:
\begin{equation}
f(\mathbf{x}) = \sgn\left( \underbrace{\sum_{i=1}^m \alpha_i k(\mathbf{x}_i, \mathbf{x})}_{\left\langle \mathbf{w}, \Phi(\mathbf{x}) \right\rangle} - \rho \right),
\end{equation}
\noindent where the offset $\rho$ can be recovered from the support vectors that lie exactly on the hyperplane, \ie, the training feature vectors whose dual variables satisfy $0 < \alpha_i < \frac{1}{\nu m}$. In this work, the projection $\left\langle \mathbf{w}, \Phi \left( \mathbf{x} \right) \right\rangle$ of a sample $\mathbf{x}$ onto the normal vector $\mathbf{w}$ is called the SVM score.

\vspace{-1mm}
\subsection{Discussion}
\label{sec:discussion}
\vspace{-1mm}

An interpretation of the solution $(\mathbf{w}^{\star}, \rho^{\star})$ for the problem stated in Eq.~\eqref{eq:ocsvm} is a hyperplane that bounds the SVM scores from below; see the inequality constraints in Eq.~\eqref{eq:ocsvm}. This interpretation also considers that the SVM score is a random variable. In this context, $\rho^{\star}$ is a threshold that discards outliers falling on the left tail of the SVM score density. Figures~\ref{fig:ocsvm_intuition}(a) and~\ref{fig:ocsvm_intuition}(b) illustrate this rationale.

However, the one-class SVM does not account for outliers that occur on the right tail of the SVM-score density. It needs to account for them to reduce false positives. Its decision rule considers these outliers as target samples yielding undesired false positives and decrease of performance.

The proposed strategy does account for these outliers. It learns two hyperplanes that tightly enclose the normal support of the SVM score density from the positive class. These hyperplanes bound the density from ``below'' and from ``above.'' The proposed strategy considers samples falling in between these hyperplanes the ``normal'' state of the positive class SVM scores. It considers samples falling outside these hyperplanes outliers: novel or abnormal samples. The region in between the hyperplanes is called a ``slab.'' In contrast with the SVM's default strategy, the proposed strategy assumes that samples from the negative class can have both negative and positive SVM scores; Figures~\ref{fig:ocsvm_intuition}(c) and~\ref{fig:ocsvm_intuition}(d) illustrate the proposed strategy.

\section{One-Class Slab Support Vector Machine}
%



This section describes the proposed one-class slab support vector machine. OCSSVM requires two hyperplanes to classify instances as negative (novel or abnormal samples) or positive (target class samples). Both hyperplanes are characterized by the same normal vector $\mathbf{w}$, and two offsets $\rho_1$ and $\rho_2$. 

The goal of OCSSVM is to find two hyperplanes that tightly enclose the region in feature space of the SVM-score density for the positive class. The positive side of each hyperplane coincides with the slab region and their negative side indicates the area where novel or abnormal samples occur; Figs.~\ref{fig:ocsvm_intuition}(c) and ~\ref{fig:ocsvm_intuition}(d) illustrate the proposed configuration of the hyperplanes and decision process.


OCSSVM solves a convex optimization problem to find the hyperplane parameters $(\mathbf{w}, \rho_1, \rho_2)$. This problem is stated as follows:
\begin{equation}
\begin{aligned}
& \underset{\mathbf{w}, \rho_1, \rho_2, \boldsymbol{\xi}, \boldsymbol{\bar{\xi}}}{\text{minimize}}
& & \frac{1}{2} \|\mathbf{w}\|^2 + \frac{1}{\nu_1 m}\sum_{i=1}^m \xi_i - \rho_1 + \frac{\varepsilon}{\nu_2 m}\sum_{i=1}^m \bar{\xi}_i + \varepsilon\rho_2 \\
& \text{subject to} 
& & \left\langle\mathbf{w}, \Phi\left(\mathbf{x}_i\right) \right\rangle \geq \rho_1 - \xi_i, \xi_i \geq 0, \\
& & & \left\langle\mathbf{w}, \Phi\left(\mathbf{x}_i\right) \right\rangle \leq \rho_2 + \bar{\xi}_i, \bar{\xi}_i \geq 0, \; i = 1, \ldots, m,
\end{aligned}
\label{eq:primal}
\end{equation}
\noindent where $(\mathbf{w}, \rho_1)$ are the parameters for the ``lower'' hyperplane $f_1$; $(\mathbf{w}, \rho_2)$ are the parameters of the ``upper'' hyperplane $f_2$, $\boldsymbol{\xi}$ and $\boldsymbol{\bar{\xi}}$ are slack variables for the lower and upper hyperplanes, respectively; $\Phi()$ is the implicit feature map in the kernel function; and $\nu_1$, $\nu_2$, and $\varepsilon$ are parameters. The parameter $\varepsilon$ controls the contribution of the slack variables $\boldsymbol{\bar{\xi}}$ and the offset $\rho_2$ to the objective function. The parameters $\nu_1$ and $\nu_2$ control the size of the slab.

This proposed optimization problem extends the formulation introduced by Sch\"{o}lkopf \etal~\cite{scholkopf_ocsvm}. It adds two new linear inequality constraints per training sample, which are the constraints for the hyperplane $f_2$, and penalty terms in the objective function of the optimization problem shown in Eq.~\eqref{eq:ocsvm}. This extension is mainly composed of linear terms and constraints. Consequently, it preserves convexity.

The offsets $\rho_1$ and $\rho_2$ have the following interpretation: they are thresholds that bound the SVM scores from the positive class (\ie, $\langle \mathbf{w}, \Phi(\mathbf{x}_i) \rangle$) from below and above, respectively. This new interpretation motivates the names for the lower and upper hyperplanes mentioned earlier. The region in between these bounds is the ``slab,'' and its size can be controlled by $\nu_1$ and $\nu_2$. The slack variables $\boldsymbol{\xi}$ and $\boldsymbol{\bar{\xi}}$ allow the OCSSVM to exclude some SVM scores that deviate from the slab region: the normal region of the SVM score density from the positive class.



The decision function of the OCSSVM, 
\begin{equation}
f(\mathbf{x}) = \sgn\left\{\left( \left\langle \mathbf{w}, \Phi\left( \mathbf{x} \right)\right\rangle - \rho_1 \right)\left(\rho_2 - \left\langle \mathbf{w}, \Phi\left( \mathbf{x} \right)\right\rangle \right) \right\},
\end{equation}
\noindent is positive when SVM scores fall inside the slab region, and negative otherwise.

Solving the primal problem (shown in Eq.~\eqref{eq:primal}) is challenging -- especially when a non-linear kernel function is used. However, the dual problem of several SVMs often yields a simpler-to-solve optimization problem. The dual problem for the OCSSVM is 

\vspace{-2mm}
\begin{equation}
\begin{aligned}
& \underset{\boldsymbol{\alpha}, \boldsymbol{\bar{\alpha}}}{\text{minimize}}
& & \frac{1}{2} \left( \boldsymbol{\alpha} - \boldsymbol{\bar{\alpha}} \right)^{T} K \left( \boldsymbol{\alpha} - \boldsymbol{\bar{\alpha}} \right) \\
& \text{subject to}
& & 0 \leq \alpha_i \leq \frac{1}{\nu_1 m}, \sum_i^m{\alpha_i} = 1,\\
& & & 0 \leq \bar{\alpha}_i \leq \frac{\varepsilon}{\nu_2 m}, \sum_i^m{\bar{\alpha}_i} = \varepsilon, \; i = 1, \ldots, m,
\end{aligned}
\label{eq:dual}
\end{equation}
\noindent where $K$ is the kernel matrix; $\alpha_i$ and $\bar{\alpha}_i$ are the $i$-th entries for the dual vectors $\boldsymbol{\alpha}$ and $\boldsymbol{\bar{\alpha}}$, respectively; and $0 \leq \nu_1 \leq 1$, $0 \leq \nu_2 \leq 1$, and $0 \leq \varepsilon$ are parameters. This dual problem is a constrained quadratic program that can be solved with convex solvers. This work considers only positive definite kernels, \ie, $K$ is positive definite~\cite{scholkopf2001learning}. Therefore,  $\varepsilon \neq 1$ must hold to avoid the trivial solution: $\boldsymbol{\alpha} = \boldsymbol{\bar{\alpha}}$.

The decision function can be re-written in terms of only the dual variables $\boldsymbol{\alpha}$, $\boldsymbol{\bar{\alpha}}$ as follows:
\vspace{-1mm}
\begin{equation}
f(\mathbf{x}) = \sgn\left\{ \left(s_{\mathbf{w}} - \rho_1 \right)\left(\rho_2 - s_{\mathbf{w}} \right)\right\},
\label{eq:decision_dual}
\end{equation}
\vspace{-1mm}
where
\vspace{-1mm}
\begin{equation}
s_{\mathbf{w}} = \left\langle \mathbf{w}, \Phi\left( \mathbf{x} \right) \right\rangle 
 = \sum_{i=1}^m \left( \alpha_i - \bar{\alpha_i}\right) k\left(\mathbf{x}, \mathbf{x}_i\right);
\label{eq:svm_score_dual}
\end{equation}
\vspace{-1mm}
and
\begin{align}
\rho_1 &= \frac{1}{N_{\text{SV}_1}} \sum_{i : 0 < \alpha_i < \frac{1}{\nu_1 m}}^{N_{\text{SV}_1}} \sum_j^m (\alpha_j - \bar{\alpha_j}) k(\mathbf{x}_i, \mathbf{x}_j) \label{eq:offset_1}\\
\rho_2 &= \frac{1}{N_{\text{SV}_2}} \sum_{i : 0 < \bar{\alpha}_i < \frac{\varepsilon}{\nu_2 m}}^{N_{\text{SV}_2}} \sum_j^m (\alpha_j - \bar{\alpha_j}) k(\mathbf{x}_i, \mathbf{x}_j) \label{eq:offset_2}.
\end{align}

The SVM score $s_{\mathbf{w}}$ is obtained from Eq.~\eqref{eq:svm_score_dual} and re-writing dot products with the kernel function. On the other hand, the offsets require analysis from the KKT conditions (see Appendix~\ref{sec:kkt_analysis}) to establish their relationship with the dual variables. The offset computation requires knowledge of the support vectors that lie exactly on the lower and upper hyperplanes. These support vectors are detected by evaluating if their dual variables satisfy $0 < \alpha_i < \frac{1}{\nu_1 m}$ and $0 < \bar{\alpha}_i < \frac{\varepsilon}{\nu_2 m}$ for the lower and upper hyperplane, respectively. Equations~\eqref{eq:offset_1} and~\eqref{eq:offset_2} require the number of support vectors $N_{\text{SV}_1}$, $N_{\text{SV}_2}$ that exactly lie on the lower and upper hyperplanes, respectively. Moreover, it can be shown via the KKT conditions that if $\alpha_i > 0$, then $\bar{\alpha_i} = 0$, and that if $\bar{\alpha_i} > 0$, then $\alpha_i = 0$. This means that each hyperplane has its own set of support vectors; the reader is referred to the Appendix~\ref{sec:kkt_analysis} for a more detailed analysis of the KKT conditions.

\vspace{-3mm}
\section{Experiments}
\label{sec:experiments}
\vspace{-2mm}

This section presents two experiments (described in Sections~\ref{sec:letter_dataset} and~\ref{sec:pascal_voc}) that assess the performance of the proposed OCSSVM. These experiments use two different publicly available datasets: the letter dataset~\cite{frey91} and the PascalVOC 2012~\cite{everingham2010pascal} dataset.

We implemented a primal-dual interior point method solver in C++\footnote{\url{http://vfragoso.com}} to find the hyperplane parameters of the proposed OCSSVM. The experiments on the letter dataset were carried out on a MacBook Pro with 16BG of RAM and an Intel core i7 CPU. The experiments on the PascalVOC 2012 dataset were executed on a machine with 32GB of RAM and an Intel core i7 CPU.

The experiments compared the proposed approach to other state-of-the-art one-class classifiers: support vector data description (SVDD)~\cite{tax2004support}, one-class kernel PCA (KPCA)~\cite{hoffmann2007kernel}, kernel density estimation (KDE), and the one-class support vector machine (OCSVM)~\cite{scholkopf_ocsvm} -- the main baseline. The experiments used the implementations from LibSVM~\cite{libsvm} for SVDD and SVM; and a publicly available Matlab implementation we created for the one-class kernel PCA algorithm to apply to the letter dataset. However, the experiments used a C++ KPCA implementation (also developed in house) for the PascalVOC 2012 dataset, since the Matlab implementation struggled with the high dimensionality of the feature vectors and large number of samples in the dataset. For the multivariate kernel density estimation, we used Ihler's publicly available Matlab toolkit~\footnote{Multivariate KDE: \url{http://www.ics.uci.edu/~ihler/code/kde.html}}. However, the KDE method did not run on the PascalVOC 2012 dataset due to the large volume of data. Thus, the experiments omit KDE results for that dataset.

The experiments trained a one-class classifier for each class in the datasets. Recall that one-class classifiers only use positive samples for training. To evaluate the performance of the one-class classifiers, the experiments used the remaining classes as negative samples (\ie, novel class instances). The tested datasets are unbalanced in this setting since there are more instances from the negative class compared to the positive class. Note that common metrics such as precision, recall, and f1-measure are sensitive to unbalanced datasets. This is because they depend on the counts of true positives, false positives, and false negatives. 


Fortunately, the Matthews correlation coefficient (MCC)~\cite{powers2011evaluation} is known to be robust to unbalanced datasets. The MCC ranges between $-1$ and $+1$. A coefficient of $+1$ corresponds to perfect prediction, $0$ corresponds to an equivalent performance of random classification, and $-1$ corresponds to a perfect disagreement between predictions and ground truth labels; see Appendix~\ref{sec:mcc} material for more details about MCC.


The experiment used common kernels (\eg, linear and radial basis function (RBF)) as well as efficient additive kernels~\cite{vedaldi2012efficient} (\eg, intersection, Hellinger, and $\chi^2$). Among these kernels, only the RBF kernel requires setting a free parameter: $\gamma$. Also, the experiment used a Gaussian kernel for the KDE method. Its bandwidth was determined by the rule-of-thumb method, an automatic algorithm for kernel bandwidth estimation included in the used Matlab KDE toolbox. The experiments compare the KDE method only with the remaining one-class classifiers using an RBF kernel since the Gaussian kernel belongs to that family.

The experiments ran a grid-search over various kernel and classifier parameters, such as $\gamma$ for the RBF kernel, $C$ parameter for SVDD, $\nu_1, \nu_2, \nu$ for the one-class SVMs, and number of components for KPCA, using a validation set for every class in every dataset; the reader is referred to the Appendix~\ref{sec:params} where these parameters are shown. 

To determine the $\varepsilon$ parameters for training the proposed OCSSVM, the experiments used a toy dataset where samples from a bivariate Normal distribution were used. It was observed that $\varepsilon = \frac{2}{3}$ produced good results; see Appendix~\ref{sec:toy_dataset} for more details of this process.

\vspace{-3mm}
\subsection{Evaluation on Letter Dataset}
\label{sec:letter_dataset}
\vspace{-2mm}

\begin{figure*}[t]
\centering
\includegraphics[width=0.85\textwidth]{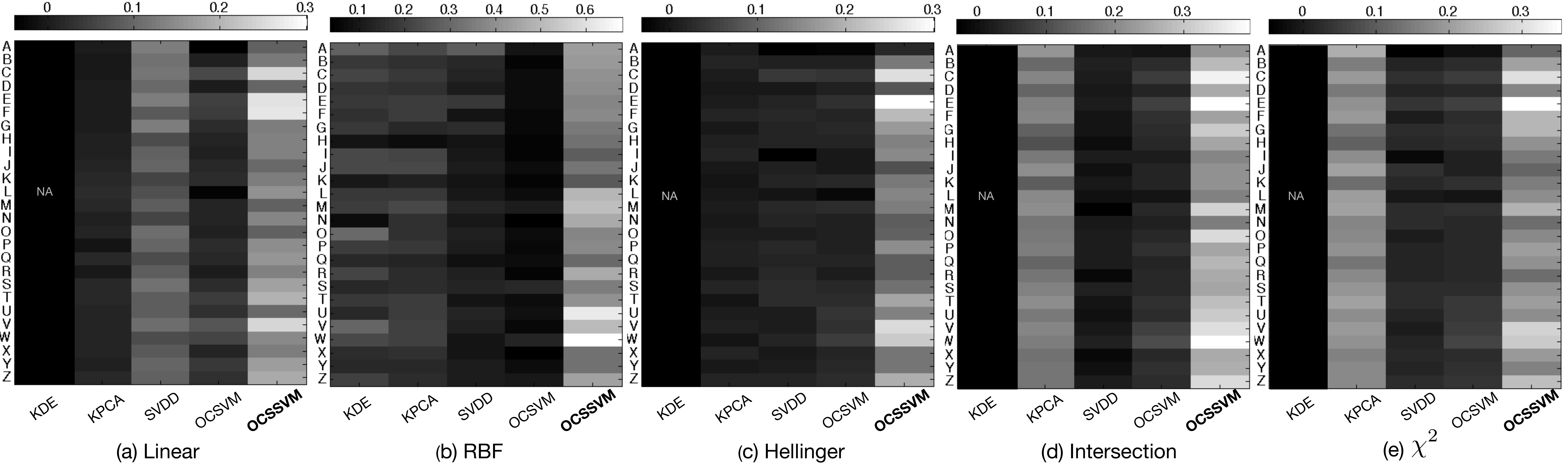}
\vspace{-3mm}
\caption{Matthews correlation coefficient on the letter dataset across different kernels (a-e); brighter indicates better performance. The proposed OCSSVM performed comparable or better than one-class kernel PCA (KPCA), kernel density estimation (KDE), support vector data description (SVDD), and one-class SVM (OCSVM). A comparison with the KDE method is only valid when using the RBF kernel.}
\label{fig:letter_mcc}
\vspace{-5mm}
\end{figure*}


\begin{table}
\caption{Median Matthews correlation coefficients over the 26 letters. Bold numbers indicate the highest score for a kernel (row). OCSSVM consistently outperformed OCSVM and performed comparable to or better than the remaining one-class classifiers.}
\vspace{-2.0mm}
\centering
\small{
\begin{tabular}{c  c  c  c  c  c  c}
\hline
Kernel & KDE & KPCA & SVDD & OCSVM & {\bf OCSSVM} \\
\hline
Linear & - & 0.01 & 0.09 & 0.02 & {\bf 0.14} \\
RBF & 0.18 & 0.17 & 0.11 & 0.07 & {\bf 0.39} \\ 
Intersection & - & 0.18 & 0.01 & 0.04 & {\bf 0.26} \\
Hellinger & - & 0.01 & 0.02 & 0.02 & {\bf 0.13} \\
$\chi^2$ & - & {\bf 0.18} & 0.02 & 0.02 & {\bf 0.18} \\
\hline
\end{tabular}
}
\label{tab:letter_average_mcc}
\vspace{-3mm}
\end{table}

This experiment aims at evaluating the performance of the OCSSVM. The tested dataset is letter~\cite{frey91}, which contains 20,000 feature vectors of the 26 capital letters in the English alphabet. Each feature vector is a 16-dimensional vector capturing statistics of a single character. The dataset provides 16,000 samples for training and 4,000 for testing. The one-class classification problem consists of training the classifier with instances of a single character (the positive class), and detecting instances of that character in the presence of novel classes -- instances of the remaining 25 characters.

Figure~\ref{fig:letter_mcc} shows the results of this experiment. It visualizes the performance of the tested classifiers across classes for different kernels. Table~\ref{tab:letter_average_mcc} presents a performance summary per kernel and per method. The results shown in Figure~\ref{fig:letter_mcc} and Table~\ref{tab:letter_average_mcc} only include a comparison of the KDE method and the one-class classifiers with an RBF kernel since the KDE method uses a Gaussian kernel, which belongs to the RBF family. Because the experiment uses Matthews correlation coefficient (MCC), higher scores imply better performance. Thus, a consistent bright vertical stripe in a visualization indicates good performance across all the classes in the dataset for a particular kernel. The figure shows that the proposed OCSSVM tends to have a consistent bright vertical stripe across different kernels and classes. This can be confirmed in Table~\ref{tab:letter_average_mcc} where OCSSVM achieves the highest median MCC for all of the kernels. The visualizations also show that the proposed OCSSVM outperformed the SVM method consistently. Comparing the OCSSVM and the SVM columns in Table~\ref{tab:letter_average_mcc} confirms the better performance of the proposed method. Table~\ref{tab:letter_average_mcc} also shows that OCSSVM performed comparable or better than one-class kernel PCA (KPCA), kernel density estimation (KDE), and support vector data description (SVDD).

\vspace{-2mm}
\subsection{Evaluation on PascalVOC 2012 Dataset}
\label{sec:pascal_voc}
\vspace{-2mm}

The goal of this experiment is to assess the performance of the  OCSSVM on a more complex dataset: PascalVOC 2012~\cite{everingham2010pascal}. This dataset contains 20 different visual classes (objects) and provides about 1,000 samples per class. It has been used mainly for object detection. The experiment used HOG~\cite{dalal2005histograms} features for every object class. To mimic novel classes that an object detector encounters, the experiment randomly picked 10,000 background regions for which HOG features were computed. The dimensionality of these features per class ranges from 2,304 to 36,864. This experiment used high-dimensional feature vectors and a large number of samples. Consequently, the kernel density estimation (KDE) MATLAB toolkit struggled and did not run properly on this dataset. Hence, the experiment omits the result for this method.

\begin{figure*}[t]
\centering
\includegraphics[width=\textwidth]{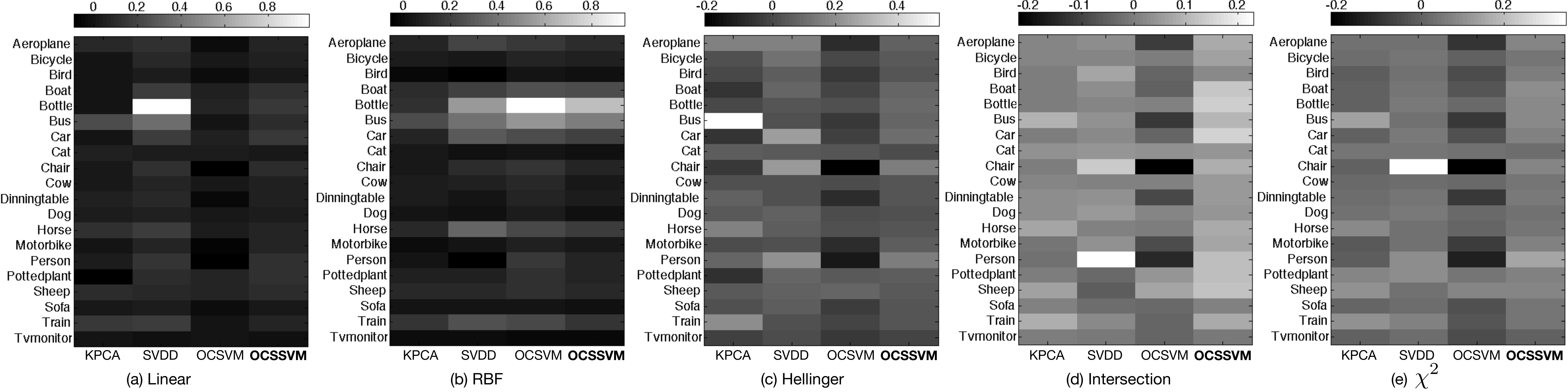}
\vspace{-7.0mm}
\caption{Average of the 3-fold Matthews correlation coefficient scores per class; brighter indicates better performance. The proposed OCSSVM outperformed the SVM using efficient additive kernels (Hellinger, Intersection, and $\chi^2$). It performed comparable or better than the one-class kernel PCA (KPCA), the support vector data description (SVDD), and the one-class SVM (OCSVM).}
\label{fig:voc}
\vspace{-5mm}
\end{figure*}

The experiment trained one-class classifiers for each object using a 3-fold cross-validation procedure. The testing set for a fold was composed of object samples and all background features. Figure~\ref{fig:voc} shows the visualizations of the average Matthews correlation coefficients (MCC) for this experiment. In addition, Table~\ref{tab:voc_average_mcc} presents a summary of this experiment.

Table~\ref{tab:voc_average_mcc} shows that OCSSVM tended to outperform the one-class SVM across kernels. Moreover, it performed comparable to or better than one-class KPCA and SVDD across kernels. Figure~\ref{fig:voc} shows that the OCSSVM tended to outperform the SVM method across classes and kernels.

\begin{table}
\caption{Median of the 3-fold Matthews correlation coefficients over the 20 classes in the PascalVOC 2012 dataset per kernel. Bold numbers indicate the highest score for a kernel (row). OCSSVM outperformed the OCSVM in most of the cases, with the exception of the RBF kernel case. It performed comparable to or better than one-class KPCA and SVDD.}
\vspace{-2.0mm}
\centering
\small{
\begin{tabular}{c  c  c  c  c}
\hline
Kernel & KPCA & SVDD & OCSVM & {\bf OCSSVM} \\
\hline
Linear & 0.02 & {\bf 0.09} & 0.01 & 0.07 \\ 
RBF & 0.05 & 0.07 & {\bf 0.14} & 0.09 \\ 
Intersection &  0.18 & 0.01 & 0.04 & {\bf 0.26} \\
Hellinger & 0.01 & 0.02 & 0.02 & {\bf 0.13} \\
$\chi^2$ & {\bf 0.18} & 0.02 & 0.02 & {\bf 0.18} \\ 
\hline
\end{tabular}
}
\label{tab:voc_average_mcc}
\vspace{-5mm}
\end{table}

\vspace{-1mm}
\section{Conclusions and Future Directions}
\vspace{-1mm}

This work presented the one-class slab support vector machine as a step towards the idealized one-class solution for open-set recognition. In contrast to the regular one-class SVM, which learns a single hyperplane for identifying target samples, instances from the positive class, the proposed classifier uses two parallel hyperplanes learned in feature space to enclose a portion of the target samples.  However, each plane has an offset with respect to the origin that places them at different locations in feature space, creating a ``slab.'' The proposed approach to train the OCSSVM is a quadratic program (QP) that estimates the hyperplane normal vector and the two offsets. 

The proposed OCSSVM showed consistent performance improvement over the regular one-class SVM on two different datasets: letter~\cite{frey91} and the PascalVOC 2012~\cite{everingham2010pascal}. The proposed strategy performed comparable or better than other state of the art one-class classifiers, such as support vector data description~\cite{tax2004support}, one-class kernel PCA~\cite{hoffmann2007kernel}, and kernel density estimation.

The approach used a Newton-based QP solver to train the OCSSVM. However, this solver is not efficient and a derivation of a sequential-minimal-optimization (SMO)~\cite{platt_smo} is planned for future work. The plan includes the adaptation of the SMO solver to deal with an extra inequality constraint that the QP of the OCSSVM includes.
\begin{appendices}



\section{KKT analysis}
\label{sec:kkt_analysis}

In this section we explore the different cases that the optimal values for the dual variables $\boldsymbol{\alpha}$ and $\boldsymbol{\bar{\alpha}}$ can fall in. As a result of this analysis, we learn how to obtain the offset values $\rho_1$ and $\rho_2$, useful conditions on support vectors for each hyperplane, and invalid cases. To do so, we exploit the KKT conditions at their optimal values. The optimal dual variables must satisfy the following statements:

\[
\left\{
\begin{array}{rl}
\alpha_i \left(\left\langle \mathbf{w}, \Phi(\mathbf{x}_i) \right\rangle - \rho_1 + \xi_i \right) & = 0 \\
\beta_i \xi_i & = 0 \\
\bar{\alpha}_i \left( \rho_2 + \bar{\xi}_i - \left\langle \mathbf{w}, \Phi(\mathbf{x}_i) \right\rangle \right) & = 0 \\
\bar{\beta}_i \bar{\xi}_i & = 0
\end{array} \right..
\]

Before starting to analyze the cases, we need to remember the following relationships:

\begin{align}
\mathbf{w} & = \sum_{i=0}^m \left( \alpha_i - \bar{\alpha}_i\right)\Phi\left( \mathbf{x} \right) \label{eq:normal}\\
\beta_i & = \frac{1}{\nu_1 m} - \alpha_i \label{eq:ineq_low}\\
1 & =\sum_i \alpha_i \label{eq:eq_low}\\
\bar{\beta}_i & = \frac{\varepsilon}{\nu_2 m} - \bar{\alpha}_i \label{eq:ineq_high}\\
\varepsilon & =\sum_i \bar{\alpha}_i \label{eq:eq_high},
\end{align}
\noindent which are obtained by differentiating the Laplacian of our problem shown in Eq. (6) of the main submission.

\subsection{Cases}

\begin{enumerate}
\item Case $\alpha_i = 0$ and $\bar{\alpha}_i = 0$. Given this scenario we conclude using Equations~\eqref{eq:ineq_low}, \eqref{eq:eq_low}, \eqref{eq:ineq_high}, and \eqref{eq:eq_high} that 

\begin{equation}
\begin{array}{rl}
\beta_i & = \frac{1}{\nu_1 m} \\
\bar{\beta}_i & = \frac{\varepsilon}{\nu_2 m}
\end{array}.
\end{equation}

\noindent Therefore,  

\begin{equation}
\xi_i = \bar{\xi}_i = 0.
\end{equation}

\noindent This implies that there are no slack variables compensating for the inequalities in the primal problem shown in Eq. (4) and thus we conclude that 

\begin{align}
\left\langle \mathbf{w}, \Phi(\mathbf{x}_i) \right\rangle & > \rho_1  \\
\left\langle \mathbf{w}, \Phi(\mathbf{x}_i) \right\rangle & < \rho_2 .
\end{align}

\noindent Samples with $\alpha_i = 0$ and $\bar{\alpha}_i = 0$ are instances that fall inside the slab.

\item Case $0 < \alpha_i < \frac{1}{\nu_1 m}$ and $\bar{\alpha}_i = 0$. In this case

\begin{equation}
\left\{ \begin{array}{rl}
\beta_i & = \frac{1}{\nu_1 m} - \alpha_i > 0 \\
\bar{\beta}_i & = \frac{\varepsilon}{\nu_2 m}
\end{array}. \right.
\end{equation}

\noindent Therefore, the following must be true

\begin{equation}
\left\{ \begin{array}{rl}
\xi_i & = 0  \\
\left\langle \mathbf{w}, \Phi(\mathbf{x}_i) \right\rangle & = \rho_1 \\
\bar{\xi}_i & = 0 \\
\left\langle \mathbf{w}, \Phi(\mathbf{x}_i) \right\rangle & < \rho_2. \\
\end{array} \right.
\end{equation}

\item Case $\alpha_i = 0$ and $0 < \bar{\alpha}_i < \frac{1}{\nu_1 m}$. In this case

\begin{equation}
\left\{ \begin{array}{rl}
\bar{\beta}_i & = \frac{\varepsilon}{\nu_2 m} - \bar{\alpha}_i > 0 \\
\beta_i & = \frac{1}{\nu_1 m}
\end{array}. \right.
\end{equation}

\noindent Therefore, the following must be true

\begin{equation}
\left\{ \begin{array}{rl}
\xi_i & = 0  \\
\left\langle \mathbf{w}, \Phi(\mathbf{x}_i) \right\rangle & = \rho_2 \\
\bar{\xi}_i & = 0. 
\end{array} \right.
\end{equation}

\item Case $0 < \bar{\alpha}_i < \frac{1}{\nu_1 m}$ and $0 < \bar{\alpha}_i < \frac{\varepsilon}{\nu_1 m}$. This implies that

\begin{equation}
\left\{ \begin{array}{rl}
\bar{\beta}_i & = \frac{\varepsilon}{\nu_2 m} - \bar{\alpha}_i > 0 \\
\beta_i & = \frac{1}{\nu_1 m} - \alpha_i > 0
\end{array}. \right.
\end{equation}

\noindent Therefore, 

\begin{equation}
\left\{ \begin{array}{rl}
\xi_i & = 0  \\
\left\langle \mathbf{w}, \Phi(\mathbf{x}_i) \right\rangle & = \rho_2 \\
\bar{\xi}_i & = 0 \\
\left\langle \mathbf{w}, \Phi(\mathbf{x}_i) \right\rangle & = \rho_1. \\
\end{array} \right.
\end{equation}

\noindent Note that this case by construction of the primal problem should not happen. This case implies that the size of the slab (\ie, $\rho_2 - \rho_1$) is zero. In other words, the two planes overlap. Therefore, there is no slab in the feature space and by construction this should not happen.

\item Case $\alpha_i = \frac{1}{\nu_1 m}$ and $\bar{\alpha}_i = 0$. This situation implies that 

\begin{equation}
\left\{ \begin{array}{rl}
\bar{\beta}_i & = \frac{\varepsilon}{\nu_2 m} \\
\beta_i & = 0
\end{array}. \right.
\end{equation}

\noindent Therefore, we conclude that

\begin{equation}
\left\{ \begin{array}{rl}
\xi_i & > 0  \\
\left\langle \mathbf{w}, \Phi(\mathbf{x}_i) \right\rangle & < \rho_2 \\
\bar{\xi}_i & = 0 \\
\left\langle \mathbf{w}, \Phi(\mathbf{x}_i) \right\rangle & < \rho_1. \\
\end{array} \right.
\end{equation}

\noindent Another implication of this case is that the $i$-th sample is considered an outlier/novel sample with respect to the first plane.

\item Case $\bar{\alpha}_i = \frac{\varepsilon}{\nu_2 m}$ and $\alpha_i = 0$. This case implies that 

\begin{equation}
\left\{ \begin{array}{rl}
\bar{\beta}_i & = 0 \\
\beta_i & = \frac{1}{\nu_1 m}
\end{array}. \right.
\end{equation}

\noindent Therefore, we conclude that 

\begin{equation}
\left\{ \begin{array}{rl}
\xi_i & = 0  \\
\left\langle \mathbf{w}, \Phi(\mathbf{x}_i) \right\rangle & > \rho_2 \\
\bar{\xi}_i & > 0 \\
\left\langle \mathbf{w}, \Phi(\mathbf{x}_i) \right\rangle & > \rho_1. \\
\end{array} \right.
\end{equation}

\noindent Again, the $i$-th sample is considered an outlier/novel sample with respect to the second plane.

\item Case $\bar{\alpha}_i = \frac{\varepsilon}{\nu_2 m}$ and $0 < \alpha_i < \frac{1}{\nu_1 m}$. In this case we have

\begin{equation}
\left\{ \begin{array}{rl}
\bar{\beta}_i & = 0 \\
\beta_i & = \frac{1}{\nu_1 m} - \alpha_i > 0 \\
\end{array}. \right.
\end{equation}

\noindent Therefore,

\begin{equation}
\left\{ \begin{array}{rl}
\xi_i & = 0  \\
\left\langle \mathbf{w}, \Phi(\mathbf{x}_i) \right\rangle & > \rho_2 \\
\bar{\xi}_i & > 0 \\
\left\langle \mathbf{w}, \Phi(\mathbf{x}_i) \right\rangle & = \rho_1. \\
\end{array} \right.
\end{equation}

\noindent This implies that $\rho_2 < \rho_1$, which again, by construction cannot happen. Thus, this case must not occur.

\item Case $\alpha_i = \frac{1}{\nu_2 m}$ and $0 < \bar{\alpha}_i < \frac{\varepsilon}{\nu_1 m}$. In this case we have

\begin{equation}
\left\{ \begin{array}{rl}
\beta_i & = 0 \\
\bar{\beta}_i & = \frac{\varepsilon}{\nu_2 m} - \bar{\alpha}_i > 0 \\
\end{array}. \right.
\end{equation}

\noindent Therefore,

\begin{equation}
\left\{ \begin{array}{rl}
\xi_i & > 0  \\
\left\langle \mathbf{w}, \Phi(\mathbf{x}_i) \right\rangle & = \rho_2 \\
\bar{\xi}_i & = 0 \\
\left\langle \mathbf{w}, \Phi(\mathbf{x}_i) \right\rangle & < \rho_1. \\
\end{array} \right.
\end{equation}

\noindent This implies that $\rho_2 < \rho_1$, which again, by construction cannot happen. Thus, this case must not occur.

\item Case $\bar{\alpha}_i = \frac{\varepsilon}{\nu_2 m}$ and $\alpha_i = \frac{1}{\nu_1 m}$. This implies that

\begin{equation}
\left\{ \begin{array}{rl}
\beta_i & \frac{1}{\nu_1 m} - \alpha_i > 0 \\
\bar{\beta}_i & = \frac{\varepsilon}{\nu_2 m} - \bar{\alpha}_i > 0 \\
\end{array}. \right.
\end{equation}

\noindent Therefore,

\begin{equation}
\left\{ \begin{array}{rl}
\xi_i & > 0  \\
\left\langle \mathbf{w}, \Phi(\mathbf{x}_i) \right\rangle & > \rho_2 \\
\bar{\xi}_i & > 0 \\
\left\langle \mathbf{w}, \Phi(\mathbf{x}_i) \right\rangle & < \rho_1. \\
\end{array} \right.
\end{equation}

\noindent This scenario implies that $\rho_2 < \rho_1$, which again, contradicts our construction of the problem. Therefore this must not occur.
\end{enumerate}

We can conclude from the analysis of these cases that any plane contains the $i$-th sample when its corresponding dual satisfies $0 < \alpha_i < \frac{1}{\nu_1 m}$ or $0 < \bar{\alpha}_i < \frac{\varepsilon}{\nu_2 m}$ for the lower and higher hyperplanes, respectively. However, only one plane can contain the $i$-th sample at a time. Therefore, at the optimal point $\alpha_i > 0$ and $\bar{\alpha_i} > 0$ does not occur. It only happens exclusively.

Thus, to recover the offsets $\rho_1$ and $\rho_2$ we need to collect all the points that satisfy either $0 < \alpha_i < \frac{1}{\nu_1 m}$ or $0 < \bar{\alpha}_i < \frac{\varepsilon}{\nu_2 m}$. Thus,

\begin{equation}
\rho_1 = \frac{1}{n_1} \sum_{i : 0 < \alpha_i < \frac{1}{\nu_1 m}} \left\langle \mathbf{w}, \Phi\left( x_i \right) \right\rangle,
\end{equation}

\noindent where $n_1$ is the number of points that satisfy $ 0 < \alpha_i < \frac{1}{\nu_1 m}$. In a similar fashion, we can recover offset $\rho_2$:

\begin{equation}
\rho_2 = \frac{1}{n_2} \sum_{i : 0 < \bar{\alpha}_i < \frac{\varepsilon}{\nu_2 m}} \left\langle \mathbf{w}, \Phi\left( x_i \right) \right\rangle,
\end{equation}

\noindent where $n_2$ is the number of points that satisfy $0 < \bar{\alpha}_i < \frac{\varepsilon}{\nu_2 m}$.

\section{Toy Dataset Experiments}
\label{sec:toy_dataset}

The goal of this experiment is twofold: 1) obtain insight about our proposed method and visualize the computed decision function for two kernels: linear and radial basis function (RBF); and 2) explore the effect of $\varepsilon$ on the learned hyperplanes. 

\subsection{Parameter Exploration}
\label{sec:epsilon}

The goal of this experiment is to determine a good value for the $\varepsilon$ parameter. To do so we generated a toy dataset composed of 1500 points drawn from a bivariate Normal distribution. We trained our one-class slab SVM using a linear kernel and an RBF kernel with $\gamma = 0.5$, with $\nu_1=0.1$ and $\nu_2=0.05$.

\begin{figure*}
\centering
\includegraphics[width=0.9\textwidth]{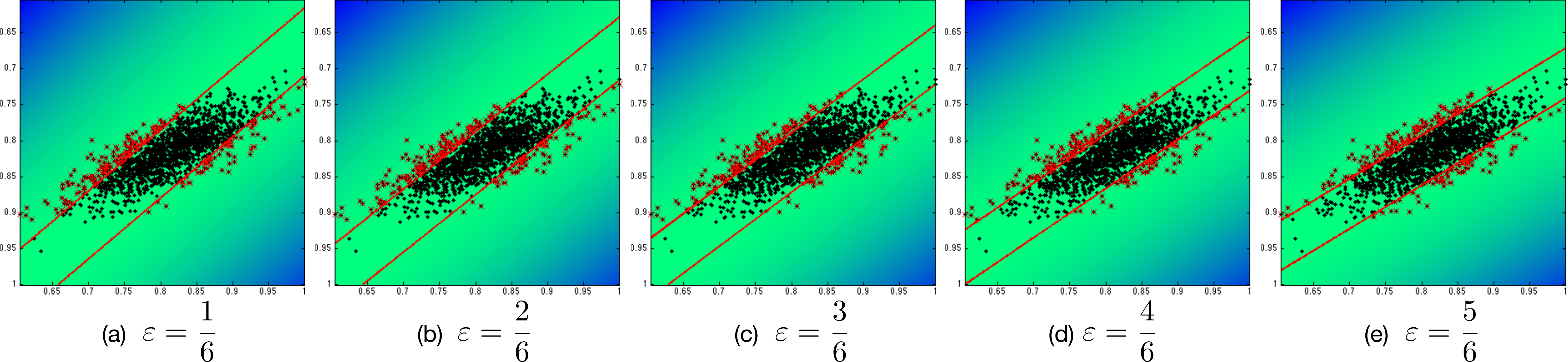}
\vspace{-3mm}
\caption{Learned hyperplanes with different $\varepsilon$ values and a linear kernel. The learned hyperplanes did not show a significant difference when varying $\varepsilon$.}
\label{fig:toy_linear}
\end{figure*}

\begin{figure*}
\centering
\includegraphics[width=0.9\textwidth]{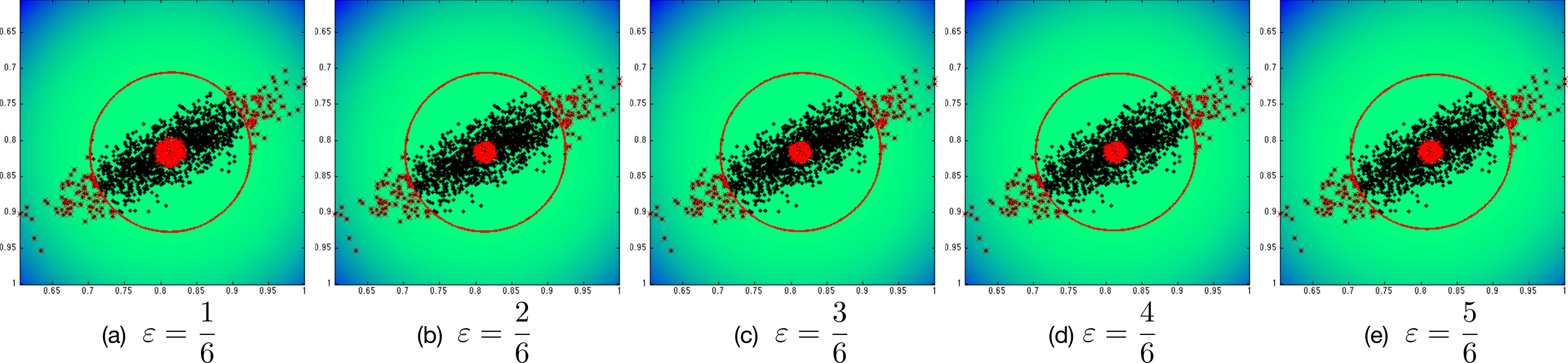}
\vspace{-3mm}
\caption{Learned hyperplanes with different $\varepsilon$ values and a RBF kernel. The learned hyperplanes did not show a significant difference when varying $\varepsilon$.}
\label{fig:toy_rbf}
\end{figure*}

\begin{table}[t]
\centering
\caption{Fraction of points that the one-class slab SVM considers as positive samples as a function of $\varepsilon$. The fraction of points labeled as positive samples did not change significantly regardless of the kernel and the value of $\varepsilon$.}
\vspace{-3mm}
\begin{tabular}{l c c c c c}
\hline
Kernel & $\varepsilon = 1/6$ & $\varepsilon = 2/6$ & $\varepsilon = 3/6$ & $\varepsilon = 4/6$ & $\varepsilon = 5/6$ \\
\hline
Linear & 0.91 & 0.91 & 0.91 & 0.91 & 0.92 \\
RBF & 0.90 & 0.90 & 0.90 & 0.90 & 0.90 \\
\hline
\end{tabular}
\label{tab:recall}
\end{table}

We varied the values of $\varepsilon$ in the interval $[\frac{1}{6}, \frac{5}{6}]$. A visualization of the hyperplanes is shown in Fig.~\ref{fig:toy_linear} and Fig.~\ref{fig:toy_rbf}. The visualizations show that there is no significant differences in the learned hyperplanes when $\varepsilon$ is varied across kernels. To verify this, we calculated the fraction of points that were considered positive by each of the learned hyperplanes. The results are shown in Table~\ref{tab:recall}. Thus we conclude that the value of $\varepsilon$ does not affect significantly the learned hyperplanes.

\subsection{Insight About One-Class Slab SVM}
\label{sec:discussion}

\begin{figure}[ht!]
\centering
\includegraphics[width=0.4\textwidth]{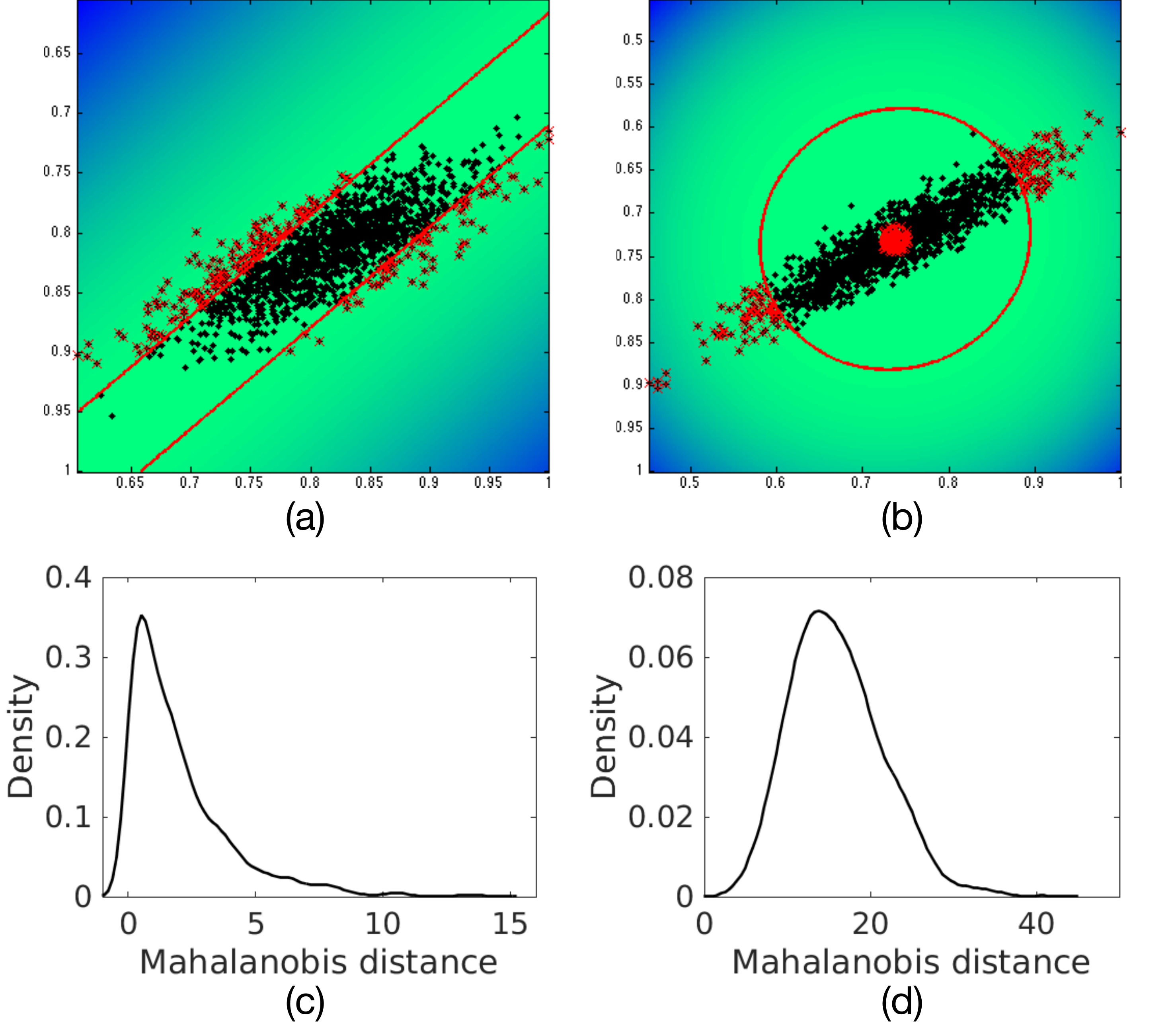}
\vspace{-4mm}
\caption{One-class slab SVM decision functions on a toy dataset. The support vectors as well as the hyperplanes are shown in red. (a) The computed slab using a linear kernel encloses most of the bivariate Normal points. (b) The ``doughnut'' like slab computed using a radial basis function (RBF) kernel captures two sets of extreme points: points deviating from the norm, and points very close to the mean. (c) The extremes found by the RBF kernel can be explained via the density of the Mahalanobis distance between the mean and a point in the dataset. It is very unlikely to observe a point very close to the mean. (d) The chances of observing a point close to the mean becomes less unlikely when the dimensionality of the points increases. This can be seen by observing the Mahalanobis distance between the mean and a point in the dataset with dimensionality 16.}
\label{fig:toy_dataset}
\end{figure}

For this experiment we set $\varepsilon = \frac{2}{3}$, $\nu_1 = 0.1$, and $\nu_2 = 0.05$. Our toy dataset is composed of 1500 points drawn from a bivariate Normal distribution. We trained our one-class slab SVM using a linear and an RBF kernel with $\gamma = 0.5$. We show a visualization of the computed decision functions in Fig.~\ref{fig:toy_dataset}.  The linear kernel finds a slab in the input space that captures most of the training data. The RBF kernel finds a slab in the input space that resembles a ``doughnut'' like slab. The RBF kernel identifies two sets of points that corresponds to the following extremes: 1) points that deviate significantly from the norm; and 2) points that fall very close to the norm. These sets of points can be verified to be ``extreme'' by analysing the density of the Mahalanobis distance between the mean and a point in the dataset. In Fig.~\ref{fig:toy_dataset}(c), not only can we observe that points falling far from the mean are rare, but also points falling very close to the mean are; the peak of the density is close to zero, but it is not exactly zero. This becomes more evident when the dimensionality of points drawn from a multivariate Normal distribution increases; see Fig.~\ref{fig:toy_dataset}(d) for an illustration.

\section{Parameters}
\label{sec:params}

In this section we present the parameters we used for the experiments presented in Section 3 of the main submission. These parameters were obtained after running a 5-fold cross validation using a validation set. The criterion was to maximize the recall rate. 

\subsection{One-class SVM parameters}
\label{sec:ocsvm_params}

\begin{table*}
\centering
\caption{RBF kernel parameter  ($\gamma$) for the letter dataset.}
\vspace{-3mm}
\begin{tabular}{c c c c c c c c c c c c c}
\hline
A & B & C & D & E & F & G & H & I & J & K & L & M \\
\hline
1.0 & 0.5 & 1.0 & 1.0 & 1.0 & 1.0 & 1.0 & 1.0 & 1.0 & 1.0 & 1.0 & 1.0 & 0.5 \\
\hline
\end{tabular}

\begin{tabular}{c c c c c c c c c c c c c}
\hline
N & O & P & Q & R & S & T & U & V & W & X & Y & Z \\
\hline
1.0 & 0.5 & 0.5 & 2.0 & 0.5 & 2.0 & 1.0 & 1.0 & 0.5 & 1.0 & 0.5 & 1.0 & 1.0 \\
\hline
\end{tabular}
\label{tab:letter_gamma_param}
\end{table*}

\begin{table}
\centering
\caption{RBF kernel parameter ($\gamma$) for the PascalVOC dataset.}
\vspace{-3mm}
\begin{tabular}{c c c c c}
\hline
Aeroplane & Bicycle & Bird & Boat & Bottle \\
\hline 
9.5367e-07 & 9.5367e-07 & 9.5367e-07 & 9.5367e-07 & 3.8147e-06 \\
\hline
\end{tabular}
\begin{tabular}{c c c c c}
\hline
Bus & Car & Cat & Chair & Cow \\
\hline
2.3842e-07 & 9.5367e-07 & 2.3842e-07 &  1.1921e-07 &9.5367e-07 \\
\hline
\end{tabular}
\begin{tabular}{c c c c c}
\hline
Diningtable & Dog & Horse & Motorbike & Person \\
4.7684e-07 & 1.1921e-07 & 9.5367e-07 & 4.7684e-07 & 1.1921e-07 \\
\hline 
\hline
\end{tabular}
\begin{tabular}{c c c c c}
\hline
Pottedplant & Sheep & Sofa & Train & Tvmonitor \\
\hline
1.9073e-06 & 1.9073e-06 & 2.3842e-07 & 9.5367e-07 & 4.7684e-07 \\
\hline
\end{tabular}
\label{tab:voc_gamma_param}
\end{table}

The $\nu$ parameter converged to $\nu=0.1$ for both datasets. The single kernel that required a parameter to be set, was the RBF kernel. For this kernel we show the parameters used for the letter and PascalVOC datasets in Table~\ref{tab:letter_gamma_param} and Table~\ref{tab:voc_gamma_param}, respectively.

\subsection{SVDD parameters}

\begin{table*}
\centering
\caption{SVDD $C$ parameter for the letter dataset.}
\vspace{-3mm}
\begin{tabular}{l c c c c c c c c c c c c c}
\hline
Kernel & A & B & C & D & E & F & G & H & I & J & K & L & M \\
\hline
Linear &  0.5 & 0.4 & 0.4 & 0.4 & 0.3 & 0.3 & 0.5 & 0.5 & 0.4 & 0.9 & 0.5 & 0.5 & 0.4 \\
RBF & 0.9 & 0.4 & 0.3 & 0.2 & 0.2 & 0.2 & 0.7 & 0.5 & 0.4 & 0.2 & 0.4 & 0.3 & 0.8 \\
Intersection & 0.1 & 0.5 & 0.9 & 0.3 & 0.1 & 0.8 & 0.6 & 0.1 & 0.6 & 0.3 & 0.7 & 0.5 & 0.5 \\
Hellinger & 0.9 & 0.6 & 0.9 & 0.1 & 0.1 & 0.1 & 0.9 & 0.1 & 0.8 & 0.3 & 0.9 & 0.2 & 0.1\\
$\chi^2$ & 0.1 & 0.1 & 0.6 & 0.4 & 0.1 & 0.1 & 0.9 & 0.1 & 0.6 & 0.4 & 0.1 & 0.1 & 0.1\\
\hline
\end{tabular}

\begin{tabular}{l c c c c c c c c c c c c c}
\hline
Kernel & N & O & P & Q & R & S & T & U & V & W & X & Y & Z \\
\hline
Linear & 0.3 & 0.3 & 0.5 & 0.5 & 0.5 & 0.5 & 0.3 & 0.5 & 0.4 & 0.5 & 0.3 & 0.4 &  0.3\\
RBF & 0.5 & 0.3 & 0.9 & 0.3 & 0.4 & 0.3 & 0.4 & 0.4 & 0.2 & 0.5 & 0.3 & 0.2 & 0.6 \\
Intersection & 0.5 & 0.7 & 0.8 & 0.2 & 0.4 & 0.8 & 0.8 & 0.1 & 0.4 & 0.1 & 0.7 & 0.4 & 0.3\\
Hellinger & 0.6 & 0.1 & 0.1 & 0.1 & 0.1 & 0.4 & 0.1 & 0.6 & 0.1 & 0.5 & 0.1 & 0.1 & 0.5 \\
$\chi^2$ & 0.8 & 0.8 & 0.9 & 0.6 & 0.1 & 0.9 & 0.6 & 0.6 & 0.1 & 0.5 & 0.1 & 0.1 & 0.8  \\
\hline
\end{tabular}
\label{tab:letter_svdd_param}
\end{table*}

The support vector data description (SVDD) method requires a parameter $C$ for training. We present the $C$ parameter we used for both experiments and per kernel in  Tables~\ref{tab:letter_svdd_param} and~\ref{tab:pascal_svdd_param}. The RBF kernel parameters used for the letter dataset and PascalVOC dataset are shown in Table~\ref{tab:letter_svdd_gamma_param} and Table~\ref{tab:voc_svdd_gamma_param}, respectively.

\begin{table*}
\centering
\caption{RBF kernel parameter ($\gamma$) for SVDD and the letter dataset.}
\vspace{-3mm}
\begin{tabular}{c c c c c c c c c c c c c}
\hline
A & B & C & D & E & F & G & H & I & J & K & L & M \\
\hline
1.0 & 0.5 & 1.0 & 1.0 & 1.0 & 1.0 & 1.0 & 1.0 & 1.0 & 1.0 & 0.5 & 1.0 & 0.5 \\
\hline
\end{tabular}

\begin{tabular}{c c c c c c c c c c c c c}
\hline
N & O & P & Q & R & S & T & U & V & W & X & Y & Z \\
\hline
0.5 & 0.5 & 0.5 & 1.0 & 0.5 & 1.0 & 1.0 & 1.0 & 0.5 & 1.0 & 0.5 & 1.0 & 0.5 \\
\hline
\end{tabular}
\label{tab:letter_svdd_gamma_param}
\end{table*}


\begin{table}[t]
\centering
\caption{SVDD $C$ parameter for the PascalVOC dataset.}
\vspace{-3mm}
\begin{tabular}{l c c c c c}
\hline
Kernel & Aeroplane & Bicycle & Bird & Boat & Bottle \\
\hline
Linear &  0.1 & 0.1 & 0.1 & 0.1 & 0.1 \\
RBF & 0.2 & 0.2 & 0.2 & 0.6 & 0.2 \\
Intersection & 0.4 & 0.1 & 0.1 & 0.2 & 0.7 \\
Hellinger & 0.1 & 0.1 & 0.7 & 0.5 & 0.7\\
$\chi^2$ & 0.1 & 0.9 & 0.2 & 0.1 & 0.9 \\
\hline
\end{tabular}

\begin{tabular}{l c c c c c}
\hline
Kernel & Bus & Car & Cat & Chair & Cow \\
\hline
Linear & 0.2 & 0.1 & 0.1 & 0.2 & 0.1 \\
RBF & 0.1 & 0.2 & 0.1 & 0.1 & 0.2 \\
Intersection & 0.2 & 0.1 & 0.4 & 0.1 & 0.4 \\
Hellinger & 0.6 & 0.1 & 0.1 & 0.9 & 0.1 \\
$\chi^2$ & 0.2 & 0.7 & 0.1 & 0.1 & 0.1 \\
\hline
\end{tabular}

\begin{tabular}{l c c c c c}
\hline
Kernel & Dinningtable & Dog & Horse & Motorbike & Person \\
\hline
Linear & 0.1 & 0.2 & 0.1 & 0.1 & 0.1 \\
RBF & 0.1 & 0.2 & 0.1 & 0.2 & 0.1 \\
Intersection & 0.4 & 0.1 & 0.3 & 0.1 & 0.6 \\
Hellinger & 0.2 & 0.1 & 0.1 & 0.1 & 0.7 \\
$\chi^2$ & 0.1 & 0.1 & 0.1 & 0.1 & 0.5\\
\hline
\end{tabular}

\begin{tabular}{l c c c c c}
\hline
Kernel & Pottedplant & Sheep & Sofa & Train & Tvmonitor \\
\hline
Linear & 0.1 & 0.2 & 0.1 & 0.1 & 0.2 \\
RBF & 0.1 & 0.1 & 0.1 & 0.1 & 0.6 \\
Intersection & 0.4 & 0.1 & 0.2 & 0.3 & 0.1 \\
Hellinger & 0.1 & 0.7 & 0.1 & 0.1 & 0.9 \\
$\chi^2$ & 0.1 & 0.1 & 0.1 & 0.1 & 0.2 \\
\hline
\end{tabular}
\label{tab:pascal_svdd_param}
\end{table}

\begin{table}[t]
\centering
\caption{RBF kernel parameter ($\gamma$) for SVDD and the PascalVOC dataset.}
\vspace{-3mm}
\begin{tabular}{c c c c c}
\hline
Aeroplane & Bicycle & Bird & Boat & Bottle \\
\hline 
4.7684e-07 & 4.7684e-07 & 4.7684e-07 & 7.6294e-06 & 7.6294e-06\\
\hline
\end{tabular}
\begin{tabular}{c c c c c}
\hline
Bus & Car & Cat & Chair & Cow \\
\hline
3.8147e-06 & 3.8147e-06 & 1.1921e-07 & 1.1921e-07 & 9.5367e-07 \\
\hline
\end{tabular}
\begin{tabular}{c c c c c}
\hline
Diningtable & Dog & Horse & Motorbike & Person \\
\hline
2.3842e-07 & 1.1921e-07 & 9.5367e-07 & 1.1921e-07 & 1.1921e-07 \\
\hline 
\end{tabular}
\begin{tabular}{c c c c c}
\hline
Pottedplant & Sheep & Sofa & Train & Tvmonitor \\
\hline
1.9073e-06 & 1.9073e-06 & 1.1921e-07 & 1.9073e-06 & 2.3842e-07 \\
\hline
\end{tabular}
\label{tab:voc_svdd_gamma_param}
\end{table}

\subsection{One-class Kernel PCA}

The number of components used in both experiments was 16. The RBF kernel parameters ($\gamma$) that we used for the letter and PascalVOC datasets are shown in Table~\ref{tab:letter_ockpca_gamma_param} and Table~\ref{tab:voc_ockpca_gamma_param}.

\begin{table*}[t]
\centering
\caption{RBF kernel parameter ($\gamma$) for one-class kernel PCA and the letter dataset.}
\vspace{-3mm}
\begin{tabular}{c c c c c c c c c c c c c}
\hline
A & B & C & D & E & F & G & H & I & J & K & L & M \\
\hline
1.0 & 0.5 & 1.0 & 1.0 & 4.0 & 2.0 & 2.0 & 16.0 & 1.0 & 4.0 & 16.0 & 4.0 & 1.0 \\
\hline
\end{tabular}

\begin{tabular}{c c c c c c c c c c c c c}
\hline
N & O & P & Q & R & S & T & U & V & W & X & Y & Z \\
\hline
0.5 & 0.5 & 2.0 & 16.0 & 4.0 & 16.0 & 2.0 & 8.0 & 16.0 & 1.0 & 2.0 & 16.0 & 16.0 \\
\hline
\end{tabular}
\label{tab:letter_ockpca_gamma_param}
\end{table*}

\begin{table}[t]
\centering
\caption{RBF kernel parameter ($\gamma$) for one-class kernel PCA and the letter dataset.}
\vspace{-3mm}
\begin{tabular}{c c c c c}
\hline
Aeroplane & Bicycle & Bird & Boat & Bottle \\
\hline
9.31E-10 & 2.38E-07 & 7.45E-09 & 9.54E-07 & 9.31E-10 \\
\hline
\end{tabular} \\
\begin{tabular}{c c c c c}
\hline
Bus & Car & Cat & Chair & Cow \\
\hline
9.31E-10 & 2.38E-07 & 9.31E-10 & 1.49E-08 & 7.45E-09 \\
\hline
\end{tabular}\\
\begin{tabular}{c c c c c}
\hline
Diningtable & Dog & Horse & Motorbike & Person \\
\hline
9.31E-10 & 9.31E-10 & 9.31E-10 & 9.31E-10 & 9.31E-10 \\
\hline 
\end{tabular}\\
\begin{tabular}{c c c c c}
\hline
Pottedplant & Sheep & Sofa & Train & Tvmonitor \\
\hline
7.63E-06 & 1.53E-05 & 9.31E-10 & 9.31E-10 & 2.38E-07 \\
\hline
\end{tabular}
\label{tab:voc_ockpca_gamma_param}
\end{table}

\subsection{One-Class Slab SVM}

The $\nu_1$ and $\nu_2$ parameters converged to $\nu_1 = 0.10$ and $\nu_2 = 0.01$ for both datasets. The single kernel that required a parameter to be set was the RBF kernel. We show the RBF parameters used for the letter and PascalVOC datasets in Table~\ref{tab:letter_ocslab_gamma_param} and Table 3~\ref{tab:voc_ocslab_gamma_param}, respectively. 


\begin{table}[t]
\centering
\caption{RBF kernel parameter  ($\gamma$) for the letter dataset.}
\vspace{-3mm}
\begin{tabular}{c c c c c c c c c c c c c}
\hline
A & B & C & D & E & F & G & H & I & J & K & L & M \\
\hline
1.0 & 0.5 & 1.0 & 1.0 & 1.0 & 1.0 & 1.0 & 1.0 & 1.0 & 1.0 & 1.0 & 1.0 & 0.5 \\
\hline
\end{tabular}

\begin{tabular}{c c c c c c c c c c c c c}
\hline
N & O & P & Q & R & S & T & U & V & W & X & Y & Z \\
\hline
1.0 & 0.5 & 0.5 & 2.0 & 0.5 & 2.0 & 1.0 & 1.0 & 0.5 & 1.0 & 0.5 & 1.0 & 1.0 \\
\hline
\end{tabular}
\label{tab:letter_ocslab_gamma_param}
\end{table}

\begin{table}[t]
\centering
\caption{RBF kernel parameter ($\gamma$) for the PascalVOC dataset.}
\vspace{-3mm}
\begin{tabular}{c c c c c}
\hline
Aeroplane & Bicycle & Bird & Boat & Bottle \\
\hline 
9.5367e-07 & 9.5367e-07 & 9.5367e-07 & 9.5367e-07 & 3.8147e-06 \\
\hline
\end{tabular}
\begin{tabular}{c c c c c}
\hline
Bus & Car & Cat & Chair & Cow \\
\hline
2.3842e-07 & 9.5367e-07 & 2.3842e-07 &  1.1921e-07 &9.5367e-07 \\
\hline
\end{tabular}
\begin{tabular}{c c c c c}
\hline
Diningtable & Dog & Horse & Motorbike & Person \\
4.7684e-07 & 1.1921e-07 & 9.5367e-07 & 4.7684e-07 & 1.1921e-07 \\
\hline 
\hline
\end{tabular}
\begin{tabular}{c c c c c}
\hline
Pottedplant & Sheep & Sofa & Train & Tvmonitor \\
\hline
1.9073e-06 & 1.9073e-06 & 2.3842e-07 & 9.5367e-07 & 4.7684e-07 \\
\hline
\end{tabular}
\label{tab:voc_ocslab_gamma_param}
\end{table}

\section{Matthews Correlation Coefficient}
\label{sec:mcc}

The MCC is computed as follows:
\begin{equation}
\text{MCC} = \frac{\text{TP} \cdot \text{TN} - \text{FN} \cdot \text{FP}}{\sqrt{(\text{TP} + \text{FP}) \cdot (\text{TP} + \text{FN})\cdot(\text{TN} + \text{FP})\cdot (\text{TN} + \text{FP})}},
\end{equation}
\noindent where the number of true positives (TP), true negatives (TN), false positives (FP), and false negatives (FN) are considered; true and false negatives are the correct and incorrect predictions of negative instances, respectively.

The MCC is positive when the product between $\text{TN} \cdot \text{TP}$ is larger than $\text{FN} \cdot \text{FP}$, which only can occur when correct predictions take place. On the other hand, it is negative when the $\text{FN} \cdot \text{FP}$ is larger than $\text{TN} \cdot \text{TP}$. The denominator ensures that the MCC metric falls in the $[-1, +1]$ range. The MCC metric is more robust for unbalanced datasets because the term measuring accurate predictions (\ie, $\text{TN} \cdot \text{TP}$) considers metrics for both classes. The MCC metric thus measures the overall accuracy of the classifier in a robust manner.
\end{appendices}

\vspace{-3mm}
\bibliographystyle{ieee}


\end{document}